\documentclass[cameraready]{Interspeech}

\title{Multilingual Word-Level Forced Alignment with Self-Supervised Representations and Learned Dynamic Programming}

\author[equalcontribution]{Roy}{Weber}
\author[equalcontribution]{Meidan}{Zehavi}
\author[equalcontribution]{Rotem}{Rousso}
\author[orcid=0000-0003-2332-5783]{Joseph}{Keshet}

\address{
    Faculty of Electrical and Computer Engineering, Technion -- Israel Institute of Technology, Haifa, 3200003, Israel
}

\email{jkeshet@technion.ac.il}


\newcommand{\bx}{\mathbf{x}}
\newcommand{\bX}{\mathbf{X}}
\newcommand{\bw}{\mathbf{w}}
\newcommand{\bs}{\mathbf{s}}
\newcommand{\ba}{\mathbf{a}}
\newcommand{\bS}{\mathbf{S}}
\newcommand{\bz}{\mathbf{z}}

\newcommand{\bU}{\mathbf{U}}
\newcommand{\R}{\mathbb{R}}

\newcommand{\V}{\mathcal{V}}

\DeclareMathOperator*{\argmax}{arg\,max}

\usepackage{comment}
\usepackage{multirow}
\usepackage[hang,flushmargin]{footmisc}

\begin{document}

\maketitle

\begin{abstract}
We present a method for accurate multilingual word-level forced alignment, consisting of an alignment encoder and a learned alignment decoder. The encoder integrates two representations: one from the Massively Multilingual Speech (MMS) model and another from a self-supervised phoneme boundary detector (UnSupSeg). It learns to fuse them and to estimate word-boundary probabilities over long temporal contexts. The alignment decoder is a learned dynamic programming that combines encoder outputs with segmental features over the MMS and UnSupSeg representations to infer final word boundaries. Trained iteratively on TIMIT and Buckeye, the proposed approach outperforms Montreal Forced Aligner (MFA) and MMS-based alignment on both datasets. On unseen languages (Dutch, German, and Hebrew), the proposed model achieves performance consistently better than or on par with existing alignment approaches, indicating its potential to scale to 1100+ languages supported by MMS without further training.
\end{abstract}



\section{Introduction}

Accurate word-level forced alignment is a fundamental component in speech and language processing. Precise temporal alignment between audio signals and textual transcriptions enables fine-grained analysis in linguistics, including the study of phonetics, phonology, prosody, and dialectal variation across languages. Beyond linguistic research, reliable alignment is critical for the development and evaluation of speech technologies such as automatic speech recognition (ASR), speech synthesis, audio indexing, and segmentation systems. 

Recent advances in sequence modeling have led to substantial improvements in speech modeling and ASR, exemplified by systems such as wav2vec 2.0~\cite{wav2vec2}, HuBERT~\cite{Hsu21-HUB}, and Whisper~\cite{Radford22-RSR}. These approaches have markedly improved recognition accuracy and robustness across varied acoustic environments and languages. However, for forced alignment tasks, traditional HMM-GMM frameworks remain widely used and highly competitive~\cite{Rousso24-FACOMP}. In particular, the Montreal Forced Aligner (MFA) \cite{McAuliffe17-MFA} has established itself as one of the leading toolkits for word- and phoneme-level alignment, consistently ranking among the top-performing systems in recent evaluations~\cite{Rousso24-FACOMP}.

The paper concludes with a comprehensive empirical evaluation. We first detail the hyperparameter tuning and model selection procedure, and then report results on multiple manually annotated speech corpora with word-level timestamps. These include TIMIT~\cite{Garofolo93-TIM}, Buckeye~\cite{Pitt05-BUC}, a Hebrew dataset~\cite{benshalom14}, the IFA~Corpus for Dutch~\cite{VanSon01-IFA}, and PHONDAT for German~\cite{tillmann1993theoretical}. The results demonstrate that our method outperforms competing approaches, including MFA and CTC-based MMS forced alignment, on the English datasets. On unseen languages, the proposed approach achieves superior performance at alignment tolerances of 50 msec and above. Code and pretrained checkpoints will be released upon acceptance\footnote{\scriptsize\url{https://github.com/MLSpeech/Multilingual-Word-Aligner}}.

\section{Method}

We assume that a waveform consisting of $T$ samples is transformed into a sequence of $L$ frames, with the frame duration of 10 msec. The speech utterance is represented as $\bX = (\bx_1, \dots, \bx_L)$, where each frame $\bx_l \in \R^d$ for $1 \le l \le L$ is a $d$-dimensional feature vector, and thus $\bX \in \R^{L \times d}$.

Let $\bw = (w_1, \dots, w_K)$ denote the sequence of words in the utterance, where $K$ is the number of words in the utterance, and each word $w_k \in \V$ belongs to the vocabulary $\V$. Our objective is to determine the start time of each word given the input speech and its corresponding word sequence. The sequence of these start times, referred to as the \emph{alignment sequence}, is denoted by $\ba = (a_1, \dots, a_K)$, where each $a_k \in \mathcal{L} = \{1, \dots, L\}$ represents the frame index at which the $k$-th word end.

Our approach builds upon $M$ pre-trained models for speech alignment representation, and aims to refine them to produce a more accurate alignment sequence than any single model alone. Each model takes as input the speech utterance $\bX$ and the corresponding word sequence $\bw$, and outputs a sequence of $L$ vectors, each indicating the confidence that the associated frame index corresponds to an alignment point. The word sequence is optional for the pre-trained models, as some of them generate the alignment sequence directly from the speech utterance alone. Specifically, the dimensionality of the vectors produced by the $m$-th model is $D_m$. The $m$-th model is then represented as $\bS_m=f_m(\bX, \bw)$ where  $\bS_m=[\bs_{m,1},\ldots,\bs_{m,L}] \in \R^{D_m \times L}$, $\bs_{m,l}\in\R^{D_m}$ for the pre-trained models $1 \le m \le M$.

The output vectors from the pre-trained models are normalized and concatenated to form the input to the alignment representation encoder, denoted $g_\theta$, where $\theta$ represents the encoder parameters. This encoder outputs $\bz=g_{\theta}(\bS)$, where $\bz\in[0,1]^L$ is a refined word boundaries given as a probability distribution over the $L$ frames, indicating the likelihood of a word boundary at each frame.

The final part of our proposed model is an alignment decoder $h_\psi$, where $\psi$ represents the decoder’s parameters. This decoder is a learned dynamic programming module that takes as input the encoder probability distribution $\bz$, the representations $\bS$, and the word sequence $\bw$ and outputs an alignment sequence corresponding to the end times of the input words, $\hat{\ba} \in \mathcal{L}^K$. 

\subsection{Alignment Representations}\label{sec:align_rep}

We used two representation models ($M=2$). The first representation model $f_1$ is based on an unsupervised phoneme segmentation framework, referred to as \emph{UnSupSeg}, which is trained using self-supervised contrastive learning \cite{Kreuk20-SCL}. This model learns to identify phoneme boundaries directly from raw speech signals without relying on annotated data, enabling it to capture fine-grained acoustic transitions that are informative for downstream word segmentation. This model takes speech samples $\bX$ as input and produces a sequence of vectors $\bS_1=[\bs_{1,1}, \ldots, \bs_{1,L}]$, where each vector $\bs_{1,l} \in \R^{D_1}$ corresponds to a 10 msec audio frame computed with an analysis window of 30 msec.

The second representation $f_2$ is derived from the MMS self-supervised speech model \cite{pratap2024scaling}. For a given speech utterance $\bX$ and a sequence of words $\bw$, CTC alignment is applied. The representation is defined as the likelihood of the $k$-th word, $w_k$, start at frame $a_k$, and it is zero otherwise. Formally,  
\begin{equation}
    s_{2,l} =
    \begin{cases}
    P(w_k \mid \bX), & l = a_k \\
    0,              & \text{otherwise}
    \end{cases}~.
\end{equation}
This representation is extracted every 20 msec and each element having dimensionality $D_2=1$. To maintain a temporal resolution of 10 ms, we upsampled the representation through simple interpolation. While alternative strategies for exploiting confidence measures derived from the CTC alignment are possible, we adopt this simple formulation as a straightforward and interpretable representation of word-level alignment confidence.

\subsection{Alignment Encoder}

The representations from the pre-trained representation models are normalized and concatenated to serve as input to the alignment encoder, whose objective is to produce a refined and improved version of these representations, $\bz=g_\theta(\bS)$ where $\bS\in \R^{L\times D}$ and $\bz\in[0,1]^L$.  We evaluated several encoder architectures for implementing $g_\theta$, including VGG \cite{Simonyan15-VGG}, Transformer encoder \cite{Hsu21-HUB}, and Conformer \cite{Gulati20-CON}. Across all architectures, the final layer was implemented as a softmax layer.

The encoder parameters $\theta$ are trained independently of the alignment decoder. Specifically, the encoder is optimized for a binary classification task that predicts whether a given frame in the input sequence corresponds to a word boundary. As boundary frames are substantially outnumbered by non-boundary frames, the task exhibits a pronounced class imbalance. To address this issue, we employ the \emph{focal loss} \cite{lin2017focal} instead of the standard cross-entropy loss, as it is explicitly designed to mitigate the effects of class imbalance. The focal loss hyperparameters ($\alpha, \gamma$) were selected via grid search on the validation set.

\subsection{Alignment Decoder}

The final component of our approach is the alignment decoder, that is a learnable dynamic programming (DP) module that refines the raw boundary probabilities produced by the encoder. Specifically, the objective of the decoder is to predict a sequence of $K$ alignments (start times) $\ba$ corresponding to the $K$ input words $\bw$. The decoder takes as input the predicted frame-level boundary probabilities $\bz = g(\bS, \bw)$, the alignment representations $\bS$, and the input word sequence $\bw$. Formally, the decoder is defined as   
\begin{equation}\label{eq:h_psi}
\hat{\ba} = \argmax_{\ba} ~ h_{\psi}(\bS, \bz, \bw, \ba).
\end{equation}
Following Keshet et al. \cite{keshet2007large}, we model the decoder as a linear combination of $N$ feature functions ${\phi_n}$ where each feature function evaluates a proposed alignment by assigning higher scores to well-placed alignments and lower scores to poorly placed ones. Formally, the decoder is formulated as follows
\begin{equation}
h_{\psi}(\bS, \bz, \bw, \ba) = \sum_{n=1}^{N} \sum_{k=1}^{K} \psi_n \phi_{n}(\bS, \bz, w_k, a_{k-1}, a_k)
\end{equation}
Each feature function captures a distinct structural aspect of alignment quality, such as temporal consistency, boundary likelihood, or agreement with the input word sequence. The feature functions used in the decoder are described next.  
Finding the alignment sequence that maximizes $h_{\psi}$ is done using a dynamic programming with constraints on the minimal word duration,
\begin{equation}
\hat{\ba} = \argmax_{\ba: a_k-a_{k-1}>L_\text{min}} ~~ \sum_{n=1}^{N} \sum_{k=1}^{K} \psi_n \phi_{n}(\bS, \bz, w_k, a_{k-1}, a_k)
\end{equation}
The parameters $\{\psi_n\}$ are optimized by iterative optimization process \cite{keshet2007large}.

The first feature function, $\phi_1$, measures the Euclidean distance between representations around a candidate boundary $a_k$. This feature is computed using the UnSupSeg model representations, under the assumption that when the boundary $a_k$ is correctly placed, the representations immediately before and after the boundary exhibit a large distance. Formally, 
\begin{equation}
  \phi_1(\bS, \bz, w_k, a_{k-1}, a_k) = \lVert \bs_{1,a_{k}-1} - \bs_{1,a_{k}+1}\rVert^2_2 ~.
  \label{equation:phi_1}
\end{equation}

 The second feature function incorporates the word transition scores derived from the encoder's output. The word transition score is calculated as follows:
\begin{equation}
  \phi_2(\bS, \bz, w_k, a_{k-1}, a_k) = z_{a_k} ~.
  \label{equation:phi_2}
\end{equation}

 The third feature function is also based on the encoder's output. For each word $w_k$, this feature function is the normalized sum of the encoder's output from the beginning of the word till the end. Namely, 
\begin{equation}
  \phi_3(\bS, \bz, w_k, a_{k-1}, a_k) = - \frac{1}{a_k{-}a_{k-1}{-}1} \sum_{l=a_{k-1}+1}^{a_k-1} z_{l} ~.
  \label{equation:phi_3}
\end{equation}
Note that this feature function is assigned a negative sign, since it yields a $\emph{low}$ value when the normalized sum of the encoder’s output corresponds to a single word, and a high value otherwise.

The fourth feature function is derived from the MMS representation \cite{pratap2024scaling}. The MMS emissions matrix, denoted as $\bU^{\text{MMS}}$, represents the probability distribution of each letter, given the speech utterance $\bX$. This matrix has dimensions equal to the number of frames and the number of letter in the language. The fourth feature function is defined as the likelihood of the emission of the characters of the word $w_k$. We sum the emissions in this time interval for the letters that make up the word:
\begin{equation}\label{equation:phi_4}
  \phi_4(\bS, \bz, w_k, a_{k-1}, a_k) = \sum_{l=a_{k-1}}^{a_k} \sum_{c \in w_{k}}   U^{\text{MMS}}_{l,c}~.
\end{equation}
This feature is high when the alignment sequence accurately represents the word sequence.

\section{Experimental Results}

\subsection{Datasets}\label{sec:datasets}

We trained and evaluated the proposed method on the TIMIT \cite{Garofolo93-TIM} and Buckeye \cite{Pitt05-BUC} speech corpora. These datasets provide manually aligned phonetic and orthographic transcriptions for read speech (5.1 hours) and conversational speech (40 hours), respectively. For each corpus, the data were partitioned at the speaker level into training, validation, and test sets using an 80/10/10 split.

We evaluate the model on TIMIT and Buckeye, and additionally on unseen languages: Hebrew, German, and Dutch. For Hebrew, we use a broadcast news dataset comprising 10 minutes of speech annotated at the phoneme level by professional linguists \cite{benshalom14}. For Dutch, we used the IFA Corpus~\cite{VanSon01-IFA}, a database of approximately five hours of hand-segmented speech from eight speakers, covering a range of speaking styles. For German, we evaluate on the full PHONDAT German corpus, which includes 201 speakers and 21,587 utterances \cite{tillmann1993theoretical}.

\subsection{Architectures, model selection, and hyperparameters}

The alignment representations are described in Sec.~\ref{sec:align_rep}. For the alignment encoder, we considered three backbone architectures: VGG \cite{Simonyan15-VGG}, a Transformer encoder \cite{Hsu21-HUB}, and Conformer \cite{Gulati20-CON}. Model selection was based on the F1-score on the validation set, reflecting the fact that the encoder predicts boundary likelihoods, which might not be the same number of input word.

Hyperparameters were tuned on the TIMIT validation set. For VGG, we evaluated depths of 13, 16, and 19 layers, with input context windows of 23, 27, 31, 35, and 41 frames. For the Transformer encoder, we considered 4, 8, and 16 layers, and context sizes of 50, 80, 100, and 200 frames. For the Conformer, we explored configurations with 12, 16, and 20 blocks, 10, 12, or 16 attention heads per block, and context windows of 250, 300, and 350 frames. The best-performing configurations were VGG19 with a 31-frame context window, a Transformer with 8 layers and a 100-frame context, and a Conformer with 16 blocks, 12 attention heads per block, a convolutional kernel size of 7, and a 300-frame context window.

\begin{table}[ht]
\centering
\caption{Word boundary prediction performance of the alignment encoder on the TIMIT and Buckeye validation sets. All metrics are computed at a frame resolution of 10 msec.}
\label{tab:sequence_models_comp}
\begin{tabular}{l | l c c c c}
\hline
\textbf{Dataset} & \textbf{Model} & \textbf{Accu.} & \textbf{Prec.} & \textbf{Recall} & \textbf{F1} \\
\hline
\multirow{3}{*}{TIMIT} 
& VGG         & 96.0 & 36.1 & \textbf{54.4} & \textbf{43.5} \\
& Transformer & 95.9 & 34.5 & 49.3 & 40.6 \\
& Conformer   & \textbf{96.5} & \textbf{40.0} & 46.5 & 43.0 \\
\hline
\multirow{3}{*}{Buckeye} 
& VGG         & 95.1 & 33.7 & 42.3 & 37.6 \\
& Transformer & 95.3 & 34.3 & 37.3 & 35.8 \\
& Conformer   & \textbf{95.3} & \textbf{35.8} & \textbf{43.0} & \textbf{39.1} \\
\hline
\end{tabular}
\end{table}

Table~\ref{tab:sequence_models_comp} reports the performance of the evaluated architectures on the TIMIT and Buckeye validation sets. The results correspond to the binary classification of word boundaries and are normalized by the relative number of input frames for each method. Performance is measured in terms of accuracy, precision, recall, and F1-score. Overall, the Conformer and VGG architectures achieved the strongest results across datasets, whereas the Transformer exhibited the lowest performance. These findings suggest that, although attention mechanisms provide a global contextual representation, the localized feature extraction enabled by convolutional components is more effective for the boundary detection task.

As the DP procedure includes a non-differentiable operation, the encoder and decoder could not be optimized jointly and were therefore trained separately. To alleviate this limitation, we fine-tuned the encoder for an additional 10 epochs, evaluating the decoder on the validation set at each epoch and applying early stopping based on validation performance. The decoder was then fine-tuned in a subsequent stage. This training procedure was applied consistently across all architectures described above. 
We evaluated alignment accuracy using the fine-tuned decoder for each model across multiple tolerance thresholds, consistent with the evaluation method in Rousso et al.~\cite{Rousso24-FACOMP}. As shown in Table~\ref{tab:models_alignment_val_comp}, both the Conformer and VGG architectures outperformed the Transformer, with comparable performance between them. However, the Conformer achieved superior results on several key metrics. 

\begin{table}[ht]
\centering
\caption{Word alignment accuracy of encoder and decoder on the TIMIT and Buckeye validation set. Thresholds are in msec.}
\resizebox{\columnwidth}{!}{%
\begin{tabular}{l | l c c c c}
\hline
\multicolumn{2}{c}{} & \multicolumn{4}{c}{\textbf{Alignment accuracy [\%]}} \\
\cmidrule(lr){3-6} 
\textbf{Dataset} & \textbf{Model} & \textbf{$t \leq 10 $} & \textbf{$t \leq 25 $} & \textbf{$t \leq 50 $} & \textbf{$t \leq 100 $} \\
\hline
\multirow{3}{*}{TIMIT} & VGG & 54.6 & \textbf{80.7} & 90.2 & 96.9 \\
                      & Transformer & 52.1 & 79.3 & 89.9 & 96.9 \\
                      & Conformer & \textbf{54.6} & 78.3 & \textbf{90.3} & \textbf{97.7} \\
\hline
\multirow{3}{*}{Buckeye} & VGG & 45.2 & 69.0 & 83.6 & \textbf{92.1} \\
                         & Transformer & 44.6 & 68.5 & 83.6 & 91.8 \\
                         & Conformer & \textbf{46.1} & \textbf{69.5} & 83.6 & 91.4 \\
\hline
\end{tabular}
}
\label{tab:models_alignment_val_comp}
\end{table}

Given its computational efficiency, robust performance across evaluation metrics, and suitability for real-time deployment, the Conformer was selected as the final model. Following this selection, the Conformer encoder was trained for 30 epochs, with early stopping based on the F1-score computed on the validation set.

\subsection{Comparison with Other Models}

Table~\ref{tab:models_alignment_test_comp} compares our approach, denoted {\bf MWA}, with established alignment systems on the TIMIT and Buckeye test sets. We evaluate against MMS CTC-based alignment \cite{pratap2024scaling}, MFA \cite{McAuliffe17-MFA}, WhisperX \cite{Bain23-WhisperX}, as reported in \cite{Rousso24-FACOMP}, and Nvidia-Canary-1B \cite{nvidia-canary}.

Our model consistently outperforms all baselines across datasets and evaluation thresholds. Notably, it surpasses MFA, a widely adopted state-of-the-art forced alignment system, as well as MMS, which serves as one of the alignment representations used in our framework. These results demonstrate that the proposed method not only improves upon existing alignment pipelines but also yields gains over the underlying representation model on which it partially builds.

\begin{table}[ht]
\centering
\caption{Word alignment accuracy of our model, MMS, MFA and WhisperX, on the TIMIT and Buckeye test set.}
\resizebox{\columnwidth}{!}{%
\begin{tabular}{l | l c c c c}
\hline
\multicolumn{2}{c}{} & \multicolumn{4}{c}{\textbf{Alignment accuracy [\%]}} \\
\cmidrule(lr){3-6}
\textbf{Dataset} & \textbf{Model} & \textbf{$t \leq 10 $} & \textbf{$t \leq 25 $} & \textbf{$t \leq 50 $} & \textbf{$t \leq 100 $} \\
\hline
\multirow{4}{*}{TIMIT} & MFA & 41.6 & 72.8 & 89.4 & 97.4 \\
                      & MMS & 18.6 & 43.5 & 75.7 & 94.7 \\
                      & WhisperX & 22.4 & 52.7 & 82.4 & 94.2 \\
                      & Nvidia-Canary-1B & 9.23 & 23.11 & 44.23 & 72.81 \\
                      & \textbf{MWA} & \textbf{58.0} & \textbf{81.3} & \textbf{91.6} & \textbf{97.8} \\
\hline
\multirow{4}{*}{Buckeye} & MFA & 39.8 & 69.9 & 84.9 & 91.8 \\
                         & MMS & 25.0 & 52.7 & 75.0 & 87.9 \\
                         & WhisperX & 18.8 & 43.1 & 67.4 & 77.4 \\
                         & Nvidia-Canary-1b & 8.06 & 18.83 & 36.31 & 63.29 \\
                         & \textbf{MWA} & \textbf{49.7} & \textbf{73.2} & \textbf{86.7} & \textbf{94.2} \\
\hline
\end{tabular}
}
\label{tab:models_alignment_test_comp}
\end{table}

\subsection{Unseen Languages}
Our alignment model was trained exclusively on English. However, since our model is based on MMS, which was pretrained on more than 1,100 languages, and on UnSupSeg, which is language-independent, we can assess the performance of the proposed system on unseen languages. Specifically, we evaluate the model's performance on Hebrew, German, and Dutch without further training. The corresponding results are reported in Table~\ref{tab:models_alignment_languages}.

We evaluate two variants of the model: one trained on TIMIT and another trained on Buckeye. Across all tested languages, the TIMIT-trained model consistently outperforms the Buckeye-trained version, suggesting that cleaner read speech provides more transferable supervision for boundary detection than conversational speech. For Hebrew, where MFA is unavailable, we compare only against MMS-based alignment. Our model outperforms MMS under stricter tolerance thresholds, while MMS performs better at more relaxed tolerances. Performance on Dutch is overall lower. On the German PHONDAT corpus \cite{tillmann1993theoretical}, the TIMIT-trained model again surpasses the Buckeye-trained model. Its performance is comparable to MFA and exceeds it at certain tolerance thresholds. Overall, these findings demonstrate that the proposed method generalizes well across languages, despite being trained exclusively on English data.

\begin{table}[h]
\centering
\caption{Word alignment accuracy of our model against MMS \cite{pratap2024scaling} and MFA \cite{McAuliffe17-MFA} on the Hebrew, German - PHONDAT, and Dutch - IFA~Corpus datasets}\label{tab:models_alignment_languages}
\resizebox{\columnwidth}{!}{%
\begin{tabular}{l | l c c c c}
\hline
\multicolumn{2}{c}{} & \multicolumn{4}{c}{\textbf{Alignment accuracy [\%]}} \\
\cmidrule(lr){3-6}
\textbf{Dataset} & \textbf{Model} & \textbf{$t \leq 10 $} & \textbf{$t \leq 25 $} & \textbf{$t \leq 50 $} & \textbf{$t \leq 100 $} \\
\hline
\multirow{2}{*}{Hebrew} & MMS & 14.3 & 41.3 & \textbf{76.5} & \textbf{94.7} \\
                      & \textbf{MWA} & \textbf{39.7} & \textbf{61.1} & 73.6 & 81.4 \\
\hline
\multirow{3}{*}{Dutch - IFA~Corpus} & MFA & 4.7 & 7.3 & 11.6 & 19 \\
                         & MMS & 16 & 37.9 & 62.9 & \textbf{76.6} \\
                         & \textbf{MWA} & \textbf{29} & \textbf{48.4} & \textbf{65.3} & 76.5 \\
\hline
\multirow{3}{*}{German - PHONDAT} & MFA & 29.9 & \textbf{65.4} & 82.1 &                                    \textbf{94.3} \\
                         & MMS & 21.8 & 44.3 & 74.9 & 91.8 \\
                         & \textbf{MWA} & \textbf{32.8} & 64.2 & \textbf{84.7} & 93.5 \\
\hline
\end{tabular}
}
\end{table}

\section{Discussion}

We proposed a method for accurate word alignment based on the MMS and an accurate self-supervised phoneme boundary representation (UnSupSeg). It does not rely on phonemes and therefore eliminates the need for G2P conversions. It proposes a potential replacement for MFA that is based on an HMM-GMM constraint model with G2P. We demonstrate the effectiveness of our approach on hand-annotated datasets in American English, Hebrew, Dutch, and German, achieving competitive and often superior performance to MFA. Since our framework relies on robust multilingual representations such as MMS, it has strong potential to scale to the full set of languages supported by MMS, enabling accurate and resource-efficient word alignment beyond English.

\section{Acknowledgments}
This work was supported by NSF DRL Grant No. 2219843 and BSF Grant No. 2022618. We also thank Rob van Son for his guidance and support with the IFA Corpus.

\section{Generative AI Use Disclosure}
Generative AI tools were used solely for language editing and manuscript polishing. They did not contribute to the scientific content, analysis, or core writing of the paper, and no AI system is listed as a co-author.

\bibliographystyle{IEEEtran}
\bibliography{references}

\end{document}